\pdfoutput=1
\documentclass[11pt]{article}

\usepackage{acl}
\usepackage{times}
\usepackage{latexsym}
\usepackage{microtype}
\usepackage{inconsolata}
\usepackage[linesnumbered,ruled,vlined]{algorithm2e}

\usepackage{amsmath,amssymb,amsfonts}
\usepackage{algorithmic}
\usepackage{graphicx}
\usepackage{textcomp}
\usepackage{xcolor}

\usepackage{cancel}
\usepackage{CJKutf8}
\usepackage{hyperref}
\usepackage{epstopdf}
\usepackage{enumitem}

\usepackage[T1]{fontenc}
\usepackage{float}
\usepackage[utf8]{inputenc}
\usepackage{multirow} 
\usepackage{makecell}
\usepackage{setspace}
\usepackage{soul}
\usepackage{subcaption}
\usepackage{booktabs}
\usepackage{tikz}

\usepackage{balance}
\definecolor{mylightgreen}{RGB}{100,200,100}
\definecolor{mylightred}{RGB}{255,66,53}
\definecolor{mylightblue}{RGB}{218,227,245} 
\definecolor{darkred}{rgb}{0.70, 0, 0}
\definecolor{darkgreen}{rgb}{0, 0.55, 0}
\definecolor{darkblue}{rgb}{0, 0.0, 0.78}
\definecolor{darkpurple}{rgb}{0.53, 0, 0.50}
\definecolor{purple}{rgb}{0.57, 0.55, 0.78}
\definecolor{iryellow}{rgb}{0.66, 0.82, 0.56}
\definecolor{trolleygrey}{rgb}{0.5, 0.5, 0.49}
\definecolor{tropicalrainforest}{rgb}{1.0, 0.91, 0.7}
\definecolor{glaucous}{rgb}{0.38, 0.51, 0.71}
\definecolor{cardinal}{rgb}{0.7, 0.33, 0.3}
\definecolor{palegreen}{rgb}{0.61, 0.7, 0.35}
\definecolor{pink}{rgb}{0.97, 0.78, 0.65}
\definecolor{orange}{rgb}{0.96, 0.69, 0.51}
\definecolor{purple}{rgb}{0.57, 0.55, 0.78}

\title{Multi-turn Natural Language to Graph Query Language Translation}

\author{
     Yuanyuan Liang$^1$, Lei Pan$^2$, Tingyu Xie$^3$, 
     Yunshi Lan$^1$, Weining Qian$^1$\\
  $^1$ East China Normal University, $^2$ AISpeech $^3$ Zhejiang University\\
  	leonyuany@stu.ecnu.edu.cn, lei.pan@aispeech.com \\
 tingyuxie@zju.edu.cn, \{wnqian, yslan\}@dase.ecnu.edu.cn 
   }

\begin{document}
\maketitle
\begin{abstract}
In recent years, research on transforming natural language into graph query language (NL2GQL) has been increasing. Most existing methods focus on single-turn transformation from NL to GQL. In practical applications, user interactions with graph databases are typically multi-turn, dynamic, and context-dependent. While single-turn methods can handle straightforward queries, more complex scenarios often require users to iteratively adjust their queries, investigate the connections between entities, or request additional details across multiple dialogue turns. Research focused on single-turn conversion fails to effectively address multi-turn dialogues and complex context dependencies. Additionally, the scarcity of high-quality multi-turn NL2GQL datasets further hinders the progress of this field. To address this challenge, we propose an automated method for constructing multi-turn NL2GQL datasets based on Large Language Models (LLMs) , and apply this method to develop the \textbf{MTGQL} dataset, which is constructed from a financial market graph database and will be publicly released for future research. Moreover, we propose three types of baseline methods to assess the effectiveness of multi-turn NL2GQL translation, thereby laying a solid foundation for future research.
\end{abstract}

\section{Introduction}
\label{sec:intro}

\begin{figure*}[ht!]
\fbox{\begin{minipage}{0.98\textwidth}
\textcolor{mylightgreen}{\textbf{User}}: Which securities stock \textcolor{darkgreen}{\textbf{opened at the highest price}} today?\\
\textcolor{blue}{\textbf{System}}: \textcolor{darkpurple}{\textbf{CITIC Securities.}}\\
\textcolor{orange}{\textbf{(GQL: match (s:stock)-[:belong\_to]->(i:industry) WHERE i.name = 'securities' return s.name order by s.\textcolor{darkgreen}{\textbf{opening\_price}} desc limit 1)}} \\
\textcolor{mylightgreen}{\textbf{User}}: What  \textcolor{darkgreen}{\textbf{ price}}?\\
\textcolor{blue}{\textbf{System}}: \textyen 30.26\\
\textcolor{orange}{\textbf{(GQL: match (s:stock \{name: '\textcolor{darkpurple}{\textbf{CITIC Securities}}'\})-[:has\_data]->(d:stock\_data \{date: '2025-01-08'\}) return d.\textcolor{darkgreen}{\textbf{opening\_price }})}} \\
\textcolor{mylightgreen}{\textbf{User}}: And yesterday?\\
\textcolor{blue}{\textbf{System}}: \textyen 36.25\\
\textcolor{orange}{\textbf{(GQL: match (s:stock \{name: '\textcolor{darkpurple}{\textbf{CITIC Securities}}'\})-[:has\_data]->(d:stock\_data \{date: '2025-01-07'\}) return d.\textcolor{darkgreen}{\textbf{opening\_price }}) }} \\
\textcolor{mylightgreen}{\textbf{User}}: How about \textcolor{purple}{\textbf{Guotai Junan}}?\\
\textcolor{blue}{\textbf{System}}: \textyen 20.00\\
\textcolor{orange}{\textbf{(GQL: match (s:stock \{name: ' \textcolor{purple}{\textbf{Guotai Junan Securities}}'\})-[:has\_data]->(d:stock\_data \{date: '2025-01-08'\}) return d.\textcolor{darkgreen}{\textbf{opening\_price }}) }} 
\end{minipage}
}
\caption{An example of a multi-turn interaction between a \textcolor{mylightgreen}{\textbf{User}} and a \textcolor{blue}{\textbf{System}}, with the orange sections representing the corresponding \textcolor{orange}{Cypher-based GQL} for each question. The color coding highlights the contextual dependencies, such as \textcolor{darkgreen}{\textbf{opening price}} , \textcolor{darkpurple}{\textbf{CITIC Securities}} and \textcolor{purple}{\textbf{Guotai Junan Securities}}.}
\label{fig:example}
\vspace{-1.0em}
\end{figure*}
% \yscomment{I suggest to highlight the dependency between the queries in a conversation with colors.}}

% 1.先引出NL2GQL
As data complexity and interconnectedness grow across various domains, graph data structures have become essential for effectively representing and analyzing relationships~\cite{zhao2022graph,sui2024unleashing}. This increasing demand for efficient data representation has driven the widespread adoption of graph databases. Consequently, graph query language (GQL) has emerged as a crucial tool for interacting with these systems, playing a pivotal role in tasks such as database management, information retrieval, and data analysis~\cite{lopes2023scalability,wang2020empirical,pavlivs2024graph}, as shown in Figure~\ref{fig:example}.
% \yscomment{give some examples of GQL such as cypher and so on.} 
However, translating natural language (NL) queries into GQL presents a significant challenge, as it requires users to possess technical expertise in database operations and a deep understanding of specific query syntax and patterns. This complexity creates a substantial barrier for individuals without a technical background~\cite{zhao2022natural,zhao2023cyspider}. To address this challenge, numerous automatic NL2GQL methods have been proposed~\cite{guo2022spcql,zhou2024r,liang2024aligning,tao2024finqa,tran2024robust}, making graph databases accessible to more audiences.
Recent advances in NL2GQL are primarily derived from the Seq2Seq framework, such as those demonstrated in~\cite{guo2022spcql} and CoBGT~\cite{tran2024robust}. With the rise of LLMs, performance has been further enhanced, leading to the development of numerous LLM-based methods~\cite{zhou2024r, liang2024aligning, tao2024finqa, liang2024nat, liu2150text}.
Alongside these methods, several NL2GQL datasets have been developed, including SpCQL~\cite{guo2022spcql}, CySpider~\cite{zhao2023cyspider}, Text2Cypher~\cite{ozsoy2024text2cypher}, $R^3$-{NL}2{GQL}\cite{zhou2024r}, TCMGQL, EduGQL\cite{liu2150text}, and StockGQL~\cite{liang2024nat}. The proposed methods and datasets mainly focus on single-turn queries.

% \yscomment{I suggest to add more description of the challenges of the multi-turn NL2GQL}.
% \yscomment{We need some sentences to describe the motivated solution to the challenges. Why LLMs? Why multi-agent?}
% ~\yscomment{Above, we did not mention the scarcity of the multi-turn NL2GQL datasets but their techniques.}

While single-turn NL2GQL translation can handle relatively simple queries, multi-turn interactions introduce several complexities that require advanced handling. First, the system must maintain context across multiple historical queries, as each new query builds upon the information provided in previous ones. This necessitates robust context management to accurately capture the user's evolving intent and ensure the generation of consistent, relevant queries. Second, as users refine or expand their queries during the interaction, the system must dynamically adjust the context to accommodate these changes. Last but not least, 
%the significant challenge lies in the scarcity of multi-turn NL2GQL datasets. 
current datasets are primarily designed for single-turn queries, resulting in limited data available for training and evaluating multi-turn systems. This constraint hampers the development of more sophisticated, context-aware solutions. 

To tackle the challenge posed by the scarcity of multi-turn NL2GQL datasets, we propose a \textbf{dependency-aware multi-turn dataset construction framework}, which performs collaborative optimization between LLMs, graph data, and dialogue dependency in an iterative way.
Our framework is composed of four essential components: a Context Manager, Question Generator, GQL Generator, and GQL Optimizer.
Here, context manager plays as a central unit to integrate the information of dialogue history and graph data and send to other constituents. Question generator, GQL generator, and GQL optimizer are LLM-based constituents to analysis the information from the context manager and output the generated questions, GQLs, and answers.
They also interact with each other for mutual checking and correction.
Using this framework, we have created the MTGQL dataset, a Chinese multi-turn NL2GQL dataset based on a financial market NebulaGraph database.

% 5.最后总结本篇论文的贡献
 Our main contributions are as follows:
\begin{itemize}[itemsep=2pt, topsep=4pt, parsep=0pt, partopsep=0pt]
    \item \textbf{A Standard Framework:} We propose a novel framework for constructing multi-turn NL2GQL datasets. To the best of our knowledge, this is the first method specifically designed for building such datasets.
    \item \textbf{MTGQL Dataset:} By applying our approach to a Chinese financial NebulaGraph database, we built MTGQL, the first Chinese multi-turn NL2GQL dataset.
    \item \textbf{Benchmark Methods:} We introduce the Backmarch methods for the MTGQL dataset, establishing a strong foundation for future research.
    
\end{itemize}

\section{Related Work}
\label{sec:rel}

\subsection{NL2GQL}
Early work in NL2GQL focused on template generation and heuristic rule-based systems. Recent advancements in NL2GQL tasks have seen a shift to deep learning-based approaches. 
Among the pioneering studies, the work~\cite{guo2022spcql} was the first to apply a Seq2Seq framework to NL2GQL, introducing a copying mechanism alongside the Seq2Seq model to enhance GQL generation. This approach paved the way for subsequent deep learning-based models in this space. The CoBGT model~\cite{tran2024robust} further advanced this field by integrating key-value extraction, relation-property prediction, and Cypher query generation. This model combines BERT, GraphSAGE, and Transformer architectures to address the NL2GQL task.

The emergence of LLMs has further advanced the research in NL2GQL.
The paper~\cite{tao2024finqa} presented a revision-based method for NL2GQL, leveraging LLMs without fine-tuning, further simplifying the process of adapting LLMs for NL2GQL tasks.
$R^3$-NL2GQL~\cite{zhou2024r} integrates small and large foundation models for ranking, rewriting, and refining tasks, enhancing query quality by better understanding context and relationships.
The work in ~\cite{liang2024aligning} proposed aligning LLMs with domain-specific graph databases to enhance query accuracy and domain relevance. It emphasizes the adaptability of LLMs when tailored to specific graph schemas, ensuring that generated queries are contextually appropriate.
In another study, ~\cite{liang2024nat} proposed a three-agent system for NL2GQL, comprising a Preprocessor for data handling, a Generator for GQL creation, and a Refiner that refines queries based on execution results. This multi-agent approach provides a more structured and efficient translation process, addressing both query generation and validation.
The method~\cite{liu2150text} proposed using template-filling and problem rewriting techniques with LLMs to provide contextual information, improving the model’s comprehension of the complex relationships between NL, graph schemas, and database data. These methods are all based on the single-turn NL2GQL task\footnote{A more detailed comparison with similar tasks is provided in the Appendix~\ref{similary_task}.}.

\subsection{NL2GQL Dataset}
The development of NL2GQL datasets has also evolved alongside advances in model architectures. Several datasets have been proposed in recent years, each addressing different aspects of the NL2GQL task. 
The SpCQL~\cite{guo2022spcql} dataset is constructed by manually annotating 10,000 NL queries with corresponding Cypher queries  based on a single Neo4j graph database.  
CySpider~\cite{zhao2023cyspider}  dataset is constructed by developing a SQL2Cypher algorithm that maps SQL queries to Cypher clauses, which are then paired with the original natural language queries to create a parallel corpus. 
Text2Cypher~\cite{ozsoy2024text2cypher} combined, cleaned, and organized several publicly available datasets into a total of 44,387 instances to enable effective fine-tuning and evaluation. 
$R^3$-NL2GQL~\cite{zhou2024r} constructed the dataset by manually creating NL-GQL pairs, using foundation models to generate diverse interpretations, and refining them manually.

Recently, using LLMs to construct data has become an effective solution to the problem of data scarcity, especially for tasks in specific domains~\cite{ding2024data,long-etal-2024-llms,zhou2024survey}.
The TCMGQL and EduGQL~\cite{liu2150text} datasets were constructed from real-world databases, ensuring standardized types and diversity. Over ten NL and GQL templates were developed based on database schema information, further enhanced by LLMs.
The work ~\cite{liang2024aligning} constructs datasets by first generating NL-GQL pairs from a graph database, followed by a two-step data augmentation process using ChatGPT to ensure diverse and comprehensive query coverage. The generated pairs are then grounded and verified.
Building upon the work in ~\cite{liang2024aligning}, the work ~\cite{liang2024nat} introduced improvements by incorporating subgraph extraction related to GQL and the colloquialization of named entities, while also constructing the StockGQL dataset. Unlike these methods, we focus on developing a multi-turn NL2GQL dataset.

\begin{figure*}[ht]
\centering
\includegraphics[width=0.98\textwidth]{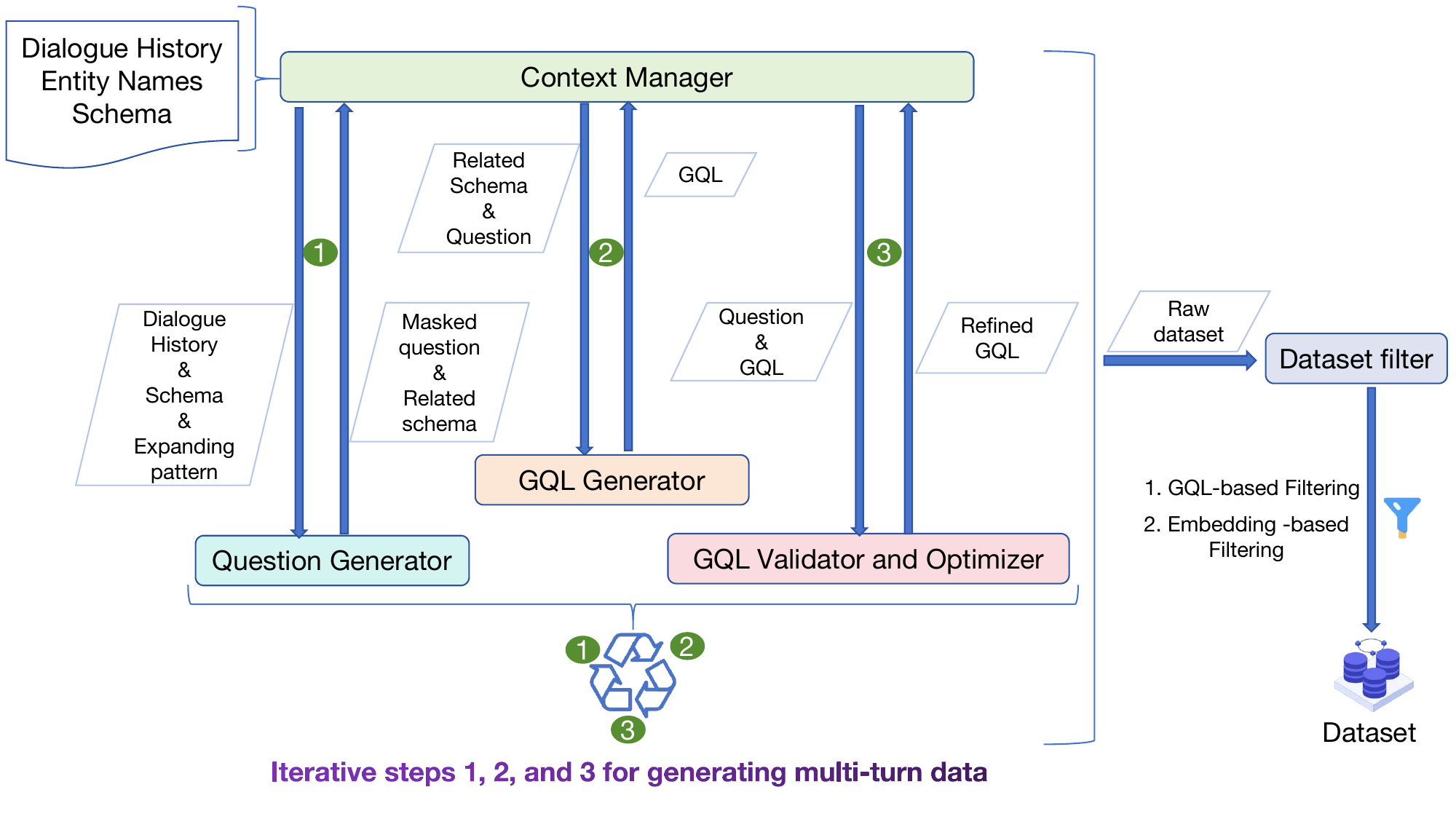}
\caption{Our framework consists of five synergistic components: the Context Manager, Question Generator, GQL Generator, GQL Validator and Optimizer, and Dataset Filter. These components work collaboratively to handle question generation, GQL generation, GQL validation and refinement, and dataset filtering. Steps 1, 2, and 3 are iteratively executed for each data point to generate multi-turn data.
}
\vspace{-1.0em}
\label{fig:Overview}
\end{figure*}

\section{Multi-turn NL2GQL Task Formulation}
\label{preliminary}
A graph database $G$ consists of a large number of connected data (nodes and edges).

We first define single-turn NL2GQL as follows. 
Given a graph database $G$ and a question $Q$, the NL2GQL system is supposed to return an executable GQL command that can be executed against $G$ and produce an answer $A$:
\[
GQL_t = \mathbb{F}(Q, G).
\]
Here, \( \mathbb{F} \) is a function that generates the graph query language \( GQL \) based on \( Q \), and \( G \).
In single-turn NL2GQL, different question-answer pairs in the dataset $\mathcal{D} = \{(Q_1, A_1), (Q_2, A_2), ...\}$ are independent.

In comparison, the interdependent question-answer pairs in multi-turn NL2GQL problem form a complete dialogue, denoted as $C = ((Q_1, A_1), (Q_2, A_2), ...,(Q_m, A_m))$ and a set of dialogues forms a dataset $\mathcal{D}= \{C_1, C_2, ...\}$.
We refer to each question-answer pair as \textit{one round of the dialogue}.
In the multi-turn NL2GQL, at the $t$-th round, given multiple rounds of historical interaction between the user $C_t$, the objective is to generate the GQL, denoted as $GQL_{t}$, corresponding to the question $Q_t$: 

\[
GQL_t = \mathbb{F}(Q_t, C_t, G),
\]
where \( C_t =\{Q_1, A_1, ..., Q_{(t-1)}, A_{(t-1)}\}\) includes all relevant user inputs and system responses executed against $G$ via the GQLs. 
%\( G \) represents the graph database, consisting of multiple nodes and edges. 

\section{A Dependency-aware Multi-turn Dataset Construction Framework}
\label{sec:data-build}

\subsection{Overview}

To construct a multi-turn NL2GQL dataset, we follow three key criteria that distinguish it from single-turn NL2GQL:
(1) \textbf{Graph Grounding:} Each question should be factually grounded via $G$ to ensure its corresponding answers can be successfully retrieved from the graph data with a GQL.
(2) \textbf{Interdependent Turns:} The question-answer pairs in a dialogue should be interdependent. Specifically, the question in the current round could be linked to the dialogue history via either questions or answers in the previous rounds.
(3) \textbf{Diverse Dependency Types:} The types of the questions and dialogue dependencies should present diversity to cover the application of practical scenario.

As showed in Figure~\ref{fig:Overview}, the framework comprises five interconnected components: \textbf{Context Manager}, \textbf{Question Generator}, \textbf{GQL Generator}, \textbf{GQL Validator and Optimizer}, and \textbf{Dataset Filter}. 
% The Context Manager serves as the core component, maintaining dialogue history, guiding data generation, selecting dependency patterns, and filling masked entities. The Question Generator ensures contextual coherence, while the GQL Generator (a fine-tuned LLM) converts questions into GQL. The GQL Optimizer then validates and corrects queries for syntax and semantics. This iterative process enables multi-turn data generation.
Next, we will detail the implementation and role of each core component.

\begin{table*}[ht!]
\centering
\resizebox{0.99\textwidth}{!}{
\begin{tabular}{l|l|l}
\toprule
 \textbf{Pattern} & \textbf{Description} & \textbf{Example} \\ \midrule
\textbf{P1: Attribute Follow-up } & \parbox{8cm}{Generates follow-up questions about an entity's attributes based on the previous query.}  &  \text{Q1: What is the largest stock in the liquor industry?} \\
 & & \text{A1: Moutai.} \\
 & & \text{Q2: What is the registered capital?} \\
\midrule

\textbf{P2: Temporal Shift } & \parbox{7cm}{Introduces the time dimension to generate queries related to historical data.} & \text{Q1: What is the highest price of Moutai today?} \\
 & & \text{A1: 20.5} \\
 & & \text{Q2: What was the closing price yesterday?} \\
\midrule
\textbf{P3: Relation Extension } & \parbox{7cm}{Expands the dialogue by querying related relationships.} & \text{Q1: What is the stock code for Tencent?} \\
 & & \text{A1: HK0700} \\
 & & \text{Q2: What is the industry data?} \\
\midrule
\textbf{P4: Same-Type Entity } & \parbox{7cm}{Used for comparative reasoning between multiple entities.} & \text{Q1: What is the opening price of Baidu today?} \\
 & & \text{A1: 150} \\
 & & \text{Q2: What about Alibaba?} \\
\midrule
\textbf{P5: Aggregation Calculation } & \parbox{7cm}{Involves queries requiring aggregation calculations such as averages or sums.} & \text{Q1: What is the opening price of Tengfei today?} \\
 & & \text{A1: 417} \\
 & & \text{Q3: What is the day-on-day growth?} \\
\midrule
\textbf{P6: Conditional Filtering } & \parbox{7cm}{Filters data based on specific conditions.} & \text{Q1: Which funds have a management fee below 1\%?} \\
 & & \text{A1: Fund A, Fund B} \\
 & & \text{Q2: Which ones have a size greater than 5 billion?} \\
\bottomrule
\end{tabular}
}
\caption{Patterns for expanding subsequent questions.}
\label{pattern_expand}
\vspace{-1.0em}
\end{table*}

\subsection{Context Manager}
\label{context_manager}
The Context Manager is  the control components of the system, Its functions include the following aspects:

\noindent \textbf{Updating the Dialogue History: } The Context Manager is responsible for maintaining the dialogue history, which includes $C_t$, the set of entities and relations, and the expansion pattern history. It continuously updates the dialogue history to ensure that all interactions are accurately tracked.

\noindent \textbf{Fulfilling Masked Questions: } Since the Question Generator generates specific entity names for certain questions but may not have access to the available entities in the database, placeholders are used. Therefore, another responsibility of the Context Manager is to replace the placeholders with actual entity names from the graph database.

\noindent \textbf{Controlling the Generation Process: } The Context Manager oversees the entire data generation process, controlling both the start and end. It is also responsible for selecting question expansion patterns based on the set of entities and relations in the history. To ensure the generation of high-quality questions, we have designed six fundamental expansion patterns, as shown in Table~\ref{pattern_expand}, and the expansion pattern selection algorithm is detailed in Appendix~\ref{pattern_expansion_algorithm}. We adjust the number of conversation rounds iteratively, keeping the total rounds per data point between 5 and 8 to maintain appropriate depth and complexity.

\subsection{Question Generator}

We use an LLM as the Question Generator, categorizing questions into initial and follow-up types. The initial question is randomly generated based on the schema of $G$, while subsequent questions follow the expansion patterns from the Context Manager. These questions must inherit context, promoting diversity, complexity, and a colloquial tone.

To better guide the LLM in generating high-quality questions, \textbf{we instruct it to produce more colloquial, informal, and ambiguous expressions that more accurately simulate real user queries.} The prompt format is shown in Appendix~\ref{generate_prompt}. It is important to note that since the Question Generator is only aware of the schema of
$G$ and does not have access to the specific entities stored within the database, questions involving entities are generated as placeholder templates. For example, \texttt{What is the opening price of [s] stock today?} where \texttt{[s]} represents a placeholder for the stock entity name. 

\subsection{GQL Generator}
The GQL Generator is responsible for generating the corresponding GQL based on the schema of $G$ and the complete question provided by the Context Manager. To enhance generation efficiency, we use the full schema to construct the prompt for fine-tuning the LLM, as outlined in Paper ~\cite{liang2024aligning}. With the fine-tuned LLM, the GQL Generator ensures accurate understanding and handling of the graph database’s schema when generating GQL.

\subsection{GQL Validator and Optimizer}
The GQL Validator and Optimizer play a crucial role in ensuring that the GQL are both syntactically and semantically correct. The workflow of the GQL Validator and Optimizer proceeds as follows: first, Syntax Validation, followed by Semantic Validation. Only GQLs containing syntax or semantic errors will undergo optimization for improvement.

\noindent \textbf{Syntax Validation:} This ensures that the generated GQL statements are syntactically correct and executable in the graph database. The GQL is executed on the database, and if it runs successfully with expected results, it is syntactically correct; otherwise, it is flagged for optimization.

\noindent \textbf{Semantic Validation:} This ensures that the GQL accurately reflects the original question's intent. We utilize the reverse generation validation method introduced in paper~\cite{liang2024aligning} to infer the original question from the generated GQL. If the vector embedding similarity between the inferred and original question is low, it indicates that the generated GQL requires further optimization.

\noindent \textbf{GQL Optimization:}  When syntax errors are detected, the system combines the original question, generated GQL, and error information into a prompt for the LLM to correct. The modified GQL is then re-validated for syntax. For semantic optimization, if the GQL doesn't align with the original question's intent, both the question and GQL are input into the LLM for correction. The corrected GQL undergoes semantic validation, and this process repeats up to three times. If all attempts fail, the system instructs the Context Manager to regenerate the question.

% \vspace{-2.5em}
\subsection{Dataset Filter}

After dataset generation, while the methods outlined above ensure the quality of each data point, they cannot guarantee the absence of similarity and redundancy. To address this, we apply two filtering methods.

\noindent \textbf{GQL-based Filtering:} We replace entity names in the GQL with placeholders and collect the masked GQL into a set. By comparing sets across data points, we calculate their similarity. If more than three identical masked GQL are found, one is discarded as redundant, effectively reducing duplicates in the dataset.

% \noindent \textbf{Embedding-based Filtering:} To prevent high similarity between questions across data points, we concatenate all questions from each entry, apply vector embedding to obtain high-dimensional representations, and calculate the similarity between data points. Any pair with similarity exceeding a preset threshold is discarded. 
% This approach effectively reduces duplicates and enhances the uniqueness, quality, and diversity of the dataset.

\noindent \textbf{Embedding-based Filtering:} To prevent high similarity between questions across data points, we concatenate all questions within each data entry and encode them using the \texttt{all-MiniLM-L6-v2} model from Sentence-BERT to obtain high-dimensional semantic representations. We then compute the cosine similarity between these vector embeddings across all data points. Any data point pair with cosine similarity exceeding a threshold of $0.6$ is considered semantically redundant and discarded. 
% This filtering strategy effectively reduces duplication and improves the uniqueness, quality, and diversity of the dataset.

Finally, we applied our approach to a Chinese financial market NebulaGraph database to develop the MTGQL dataset based on nGQL syntax.

\section{Data Analysis}
\label{sec:analysis}

% \yscomment{Please describe the statistics and evaluation of the data set.}

\subsection{Dataset Statistics}

As shown in Table~\ref{chapter6_static_result}, the dataset contains 4,500 multi-turn dialogues, split into 3,000 for training, 500 for development, and 1,000 for testing. Each dialogue has an average of 6.49 turns, reflecting balanced dialogue depth. In total, there are 29,196 GQL statements, with multiple queries per dialogue, indicating the dataset’s complexity. On average, each dialogue involves 4.79 entities and 5.59 relations, requiring models to handle rich and diverse graph structures. The slightly higher averages in the test set suggest a more challenging evaluation setting. Overall, the dataset is well-structured and suitable for training and evaluating models on dialogue-based graph query tasks.

\subsection{Human Evaluation}

We evaluated the quality of the dataset by asking three domain experts to rate 200 randomly selected dialogues from each of the training, validation, and test sets. The evaluation focused on four dimensions: coherence, question diversity, coverage, and semantic accuracy, using a 1–5 scale. As shown in Table~\ref{human_result}, the results confirm the dataset’s effectiveness for training and evaluating dialogue systems. Additionally, we recalculated Cohen’s Kappa and obtained a score of 85.76, indicating a high level of inter-rater agreement. More information on the manual evaluation can be found in Appendix~\ref{man_evlu_pro}.
\begin{table}[ht!]
\centering
\tabcolsep=0.4em
\renewcommand\arraystretch{0.8}
\begin{tabular}{llll}
\toprule
                                   & train  & dev   & test  \\ \midrule
Coherence                    & 4.48   & 4.31   & 4.17   \\ \midrule
Question Diversity & 4.16   & 4.08   & 4.01   \\ \midrule
Semantic Accuracy        & 4.68   & 4.52   & 4.38   \\  \bottomrule
\end{tabular}
\caption{Human evaluation results.}
\label{human_result}
\vspace{-1.5em}
\end{table}

\subsection{Comparison with Other Datasets}

As shown in Table~\ref{summary_datasets}, the table compares several NL2GQL datasets, with MTGQL standing out as the only multi-turn dataset. Unlike other single-turn datasets, MTGQL is specifically designed to handle more complex, multi-turn queries, making it particularly suitable for tasks that require multiple interactions. Therefore, MTGQL will play a pivotal role in advancing research in multi-turn NL2GQL. 
For a more detailed description of the dataset generation methodology and dataset analysis, please refer to Appendix~\ref{method_dataset_analyse}.
% As shown in Table~\ref{summary_datasets}, MTGQL is the only multi-turn NL2GQL dataset, distinguishing it from other single-turn datasets. Specifically designed for complex, multi-turn queries, MTGQL is essential for advancing research in multi-turn NL2GQL tasks.

\begin{table*}[ht!]
\centering
% \resizebox{0.98\textwidth}{!}{
\tabcolsep=1.1em
\renewcommand\arraystretch{0.8}
\begin{tabular}{lllll}
\toprule
                                   & train  & dev   & test  & total  \\ \midrule
Number of Data Points         & 3000   & 500   & 1000  & 4500   \\ \midrule
Total Number of GQLs           & 19320 & 3252 & 6624 & 29196 \\ \midrule
Average Dialogue Turns per Data & 6.44  & 6.50  &  6.62 & 6.49    \\  \midrule
Average entity per Data          & 4.64  & 4.89  & 5.17  &  4.79  \\ \midrule
Average relation per Data          & 5.47  & 5.65  & 5.93  &  5.59 \\ 
\bottomrule
\end{tabular}
% }
\caption{Basic Statistics of the Dataset.}
\label{chapter6_static_result}
\vspace{-0.5em}
\end{table*}
\begin{table*}[ht!]
\centering
% \resizebox{0.99\textwidth}{!}{
\tabcolsep=0.5em
\renewcommand\arraystretch{0.8}
\begin{tabular}{lccccc}
\toprule
Dataset      & Language                                                  & Multi or Single  & Domain      & Syntax & Number \\ \midrule
SpCQL~\cite{guo2022spcql}        & Chinese                                                   & Single                & Open-domain & Cypher & 10000  \\ \midrule
CySpider~\cite{zhao2023cyspider}  & English                                                   & Single                & Open-domain & Cypher & 4929   \\ \midrule
Text2Cypher~\cite{ozsoy2024text2cypher}   & English                                                   & Single                & Open-domain & Cypher & 44387 \\ \midrule

FinGQL~\cite{liang2024aligning}   & Chinese                                                   & Single                & Finance & nGQL & - \\ \midrule
MediGQL~\cite{liang2024aligning}   & Chinese                                                   & Single                & Medicine & Cypher & - \\ \midrule
$R^3$-NL2GQL~\cite{zhou2024r}  & \begin{tabular}[c]{@{}c@{}}Chinese\\ English\end{tabular} & Single                & Open-domain & nGQL   & -   \\ \midrule
StockGQL~\cite{liang2024nat}      & Chinese                                                   & Single                & Stock       & nGQL   & 6456   \\ \midrule
TCMGQL~\cite{liu2150text}        & Chinese                                                   & Single                & Medicine    & Cypher & -      \\ \midrule
EduGQL ~\cite{ liu2150text}       & Chinese                                                   & Single                & Education   & Cypher & -      \\ \midrule
\textbf{MTGQL(Ours) }      & Chinese                                                   & \textbf{Multi}                 & Stock       & nGQL   & 4500   \\ \bottomrule
\end{tabular}
% }
\caption{A summary of the main NL2GQL datasets. From this, we can conclude that MTGQL is the only multi-turn dataset. The "-" in the Number column indicates that the dataset has not been open-sourced yet.}
\label{summary_datasets}
\vspace{-1.0em}
\end{table*}

\section{Models and Experimental Setup}

\subsection{Benchmark Methods}

\noindent \textbf{In-context learning with all schema method (ICL-AS):} This method provides a set of examples within the input prompt, which concatenates all schema information and the question, guiding the LLM to generate the corresponding GQL. 

\noindent \textbf{Related schema extraction method (RSE):} During training, this method uses the related schema and question as input, with the labeled GQL as output, while fine-tuning the LLM. In inference, it guides the LLM to extract related schema. 

\noindent \textbf{Fine-tuning with with all schema method (FT-AS):} Approach concatenates all schema information with the question as input while applying LoRA for parameter-efficient fine-tuning of the base LLM. 

\noindent \textbf{Dependency-aware method (DA):} 
We adapt the Dependency-aware Multi-turn Dataset Construction Framework with minor modifications and follow the method proposed in~\cite{liang2024nat} to construct a dependency-aware baseline. The adapted method comprises three key modules: a \textit{Context Manager}, a \textit{GQL Generator}, and a \textit{GQL Refiner}.
First, the \textbf{Context Manager} maintains the dialogue history, including previous questions, corresponding GQL queries and answers, as well as the involved entities and relations. It reformulates the current question based on the dialogue history to make it more formal and information-rich. Additionally, it extracts the relevant sub-schema for the current turn.
Second, the \textbf{GQL Generator} generates a GQL query based on the reformulated question and the extracted sub-schema.
Third, the \textbf{GQL Refiner} improves the generated query by refining it based on its execution results to enhance accuracy and relevance.
More details are provided in Appendix~\ref{da_method}.

\begin{table*}[ht!]
\centering
% \resizebox{0.99\textwidth}{!}{
\tabcolsep=0.6em
\renewcommand\arraystretch{0.8}
\begin{tabular}{l l c c c c }
\toprule
\multirow{1}{*}{\textbf{Method}} & \textbf{Backbones}& \textbf{EM(\%)} & \textbf{AEM(\%)} & \textbf{EX(\%)} & \textbf{AEX(\%)} \\
\midrule
\multirow{4}{*}{ICL-AS}
& GLM-4-9B-Chat         &31.13    & 6.50   & 30.01  & 5.80   \\
& LLaMA-3.1-8B-Instruct & 27.66   & 6.10   & 27.76  & 6.40 \\
& Qwen2.5-14B-Instruct  & 32.55   & 7.50   & 29.70  & 7.20  \\
& ChatGPT-4o            &  38.29  & 10.9  &  36.28  & 8.80\\
\midrule
\multirow{2}{*}{RES }  
&GLM-4-9B-Chat          & 56.91  & 25.70  & 53.64  &  22.30   \\
& LLaMA-3.1-8B-Instruct &  58.76 & 27.10  & 56.63  & 26.70 \\
& Qwen2.5-14B-Instruct  & 59.60  & 28.30  & 57.71  &  26.80 \\ 

\midrule
\multirow{2}{*}{FT-AS}  
& GLM-4-9B-Chat         & 60.14  & 30.60 & 56.16  & 28.80  \\  
& LLaMA-3.1-8B-Instruct & 61.23  & 31.10 & 60.19  & 29.20 \\
& Qwen2.5-14B-Instruct  & 63.56  & 31.50 & 61.70  & 31.20    \\ 
\midrule
\multirow{2}{*}{DA}  
& GLM-4-9B-Chat         & 65.53  & 38.70  & 63.47  & 36.60  \\  
& LLaMA-3.1-8B-Instruct & 66.73  & 38.40  & 63.36  &  37.20 \\
& Qwen2.5-14B-Instruct  & \textbf{68.45}  &\textbf{40.60 } & \textbf{65.39}  &  \textbf{38.30 }   \\ 
\bottomrule
\end{tabular}
% }
\caption{\label{table:main_result}
The comparison between the baseline methods is shown, with the bold numbers indicating the best results.
}
\vspace{-1.0em}
\end{table*}

\subsection{Experimental Setup}

\noindent \textbf{Evaluation Metrics.}
The work in ~\cite{guo2022spcql} introduced Exact Match (EM) and Exact Explanation (EX) for single-turn tasks. For multi-turn tasks, we propose Overall Exact Match (AEM) and Overall Exact Explanation (AEX), where all turns in a dialogue must be correct for the data to be considered successful. The formulas are as follows:
\setlength{\abovedisplayskip}{0pt}
\setlength{\belowdisplayskip}{0pt}
\begin{equation}
EM = \frac{\substack{\text{number of GQLs with a correct logical form} }}{\text{total number of GQL}}
\end{equation}

\begin{equation}
\text{AEM} = \frac{\substack{\text{number of data points with all GQLs } \\ \text{ having correct logical form}}}{\text{total number of data points}}
\end{equation}

\begin{equation}
EX = \frac{\substack{\text{number of GQLs with a correct execution result}}}{\text{total number of GQL}}
\end{equation}

\begin{equation}
\text{AEX} = \frac{\substack{\text{number of data points with all GQLs } \\ \text{having correct execution results}}}{\text{total number of data points}}
\end{equation}
\setlength{\abovedisplayskip}{12pt}  % 默认值，可能略有不同
\setlength{\belowdisplayskip}{12pt}

\noindent \textbf{Implementation Details.}
Our experiments were conducted on an A800 GPU. We selected Qwen2.5-14B-Instruct~\cite{qwen2.5}, LLaMA-3.1-8B-Instruct~\cite{dubey2024llama}, and GLM-4-9B-Chat~\cite{glm2024chatglm} as the LLM backbone models. In this paper, all sequence encoding is performed using the all-MiniLM-L6-v2 model, with the embedding dimension set to 384. All the number of demonstrations $K$ are set as $4$.

\section{Results}
% In the following sections, we highlight our key findings and results. For comprehensive results and detailed tables, please refer to the Appendix.

\subsection{Main Results}
Based on the results presented in Table~\ref{table:main_result}, the DA method consistently outperforms all other approaches across all evaluation metrics. Notably, when combined with the Qwen2.5-14B-Instruct backbone, DA achieves the highest scores in EM (\textbf{68.45\%}), AEM (\textbf{40.60\%}), EX (\textbf{65.39\%}), and AEX (\textbf{38.30\%}).
In contrast, the ICL-AS method yields comparatively lower results, which can be attributed to the absence of high-quality GQL-related corpora during the pretraining of its underlying models.
Moreover, performance differences observed across various backbone models within the same method underscore the substantial impact of model architecture and backbone selection on the final outcomes. This highlights the necessity of carefully choosing and aligning the model backbone with the specific demands of the task.
Nevertheless, it is worth noting that the overall accuracy on this task remains relatively low, suggesting that there is still considerable room for improvement.

% \yscomment{A systematic evaluation of different models on the proposed dataset.}

\subsection{Breakdown Results by Round}
Table~\ref{breakdown_result_round} presents the results of the best baseline method across different rounds, showing a clear decline in performance as rounds increase. 
% The accuracy drops from 91.20\% in Round 1 (R1) to 80.28\% in Round 5 and beyond (R5+). 
This decrease is likely due to the increasing complexity of multi-turn interactions, which challenges the model's ability to maintain context and generate consistent responses.

\begin{table}[ht!]
\centering
\small
% \resizebox{0.45\textwidth}{!}{
\tabcolsep=0.8em
\renewcommand\arraystretch{0.2}
\begin{tabular}{l  c c }
\toprule
\textbf{Round} & \textbf{EM(\%)} &   \textbf{EX(\%)}  \\
\midrule

R1       & 84.21  &  82.88 \\ \midrule
R2       &  73.66  &  73.13  \\ \midrule
R3       &  60.25 &  58.44 \\ \midrule
R4       &  47.84 &  46.18 \\ \midrule
R5+       & 31.23  &  30.96  \\ \bottomrule
\end{tabular}
% }
\caption{\label{breakdown_result_round}
The breakdown of results by round, where R1-R4 represent rounds 1 to 4, and R5+ denotes round 5 and beyond.
}
\vspace{-1.0em}
\end{table}

\begin{table}[ht!]
\centering
\small
% \resizebox{0.45\textwidth}{!}{
\tabcolsep=0.8em
\renewcommand\arraystretch{0.2}
\begin{tabular}{l  c c }
\toprule
\textbf{Round} & \textbf{EM(\%)} &  \textbf{EX(\%)}  \\
\midrule
P1       & 70.47  &  68.49  \\ \midrule
P2       &  64.70 & 63.66\\ \midrule
P3       & 66.52  & 64.12   \\ \midrule
P4       & 73.84  & 71.68  \\ \midrule
P5       & 62.59  &  62.32 \\ \midrule
P6       & 67.36  & 66.46  \\ \bottomrule
\end{tabular}
% }
\caption{\label{breakdown_result_type}
Results by the question expansion pattern.
}
\vspace{-1.5em}
\end{table}

Table~\ref{breakdown_result_type} shows performance across different question expansion patterns, with notable variations. 
% For example, P1 achieves an EM of 89.26\%, while P5 drops to 80.47\%. 
These fluctuations indicate that the model is more effective with simpler question expansions (like P1 and P4), while more complex patterns (like P2 and P5) lead to lower accuracy, likely due to the increased difficulty of generating precise answers. More experimental analyses are provided in Appendix~\ref{further_expr}.

\section{Conclusion}
\label{sec:econ}

In this paper, we introduce a dependency-aware multi-turn dataset construction framework for building multi-turn NL2GQL datasets. Using this framework, we create MTGQL, the first multi-turn NL2GQL dataset. Finally, we propose three baseline methods based on this dataset, laying the groundwork for future advancements in the field.
\section*{Limitations}     
There are several limitations that we would like to address in future work.

First, although we have developed a Chinese multi-turn NL2GQL dataset, we have not yet completed the translation into English due to the extensive amount of entity and relation names that require translation from the graph database. Once this process is completed, we plan to release a bilingual (Chinese-English) version of the dataset as open source to facilitate broader research adoption.

Second, while our dataset supports multi-turn queries involving complex contextual dependencies, the current benchmark methods rely on manually designed schemas or dependency-aware modules. These methods may not generalize well to unseen domains or schema structures. Future work could explore schema-agnostic approaches or large-scale pretraining on multi-turn graph querying tasks.

Third, the current evaluation focuses primarily on execution accuracy of generated GQL. However, execution correctness may not fully capture semantic correctness or partial matching of subgraph intents. Incorporating human evaluation or developing more fine-grained metrics could provide better insights into model behavior.

Lastly, although our dataset construction process includes context reformulation and sub-schema extraction, the pipeline still involves certain heuristic rules and prompt designs that may not scale well across diverse graph domains. We aim to further automate and generalize the dataset construction framework to reduce reliance on manual tuning.

\bibliography{custom}

% \appendix
\clearpage
\section{Appendix}
\subsection{Comparison with Similar Tasks }
\label{similary_task}

\noindent \textbf{Text2SQL}

While numerous highly effective Text2SQL methods have been developed~\cite{caferouglu2024sql,wang2023mac,talaei2024chess}, the fundamental differences between GQL and SQL present significant challenges for directly applying these methods to the NL2GQL task. Several studies have examined the differences between Text2SQL and NL2GQL~\cite{guo2022spcql,liang2024aligning,zhou2024r}, and we highlight the key distinctions in the following areas:

\begin{itemize}
    \item \textbf{Differences in standard syntax:} Unlike SQL, which follows a standardized query language, GQL lacks a unified standard. Different graph databases adopt distinct query languages such as Cypher, nGQL, and Gremlin. This fragmentation complicates dataset construction, model generalization, and the development of consistent training paradigms.
    \item \textbf{Differences in query types:} GQL surpasses the typical CRUD operations by offering advanced query types like sub-graph and path queries that enable complex data traversal. Its extensive keyword set further enhances its flexibility, making it a powerful tool for a wide range of data manipulation needs.
    \item \textbf{Differences in translation difficulties:} NL2GQL involves understanding graph structures, path reasoning, and pattern matching, requiring high query flexibility, which may lead to issues such as path combination explosion. In contrast, Text-to-SQL faces challenges like pattern matching, table/column name mapping, and SQL syntax parsing, but the overall query structure remains relatively stable.
    \item \textbf{Differences in language model capabilities:} Text-to-SQL benefits from a large corpus and extensive datasets, while NL2GQL has far fewer resources. Given that most widely used pre-trained models, especially LLMs, rely on pre-training followed by fine-tuning, this disparity in resources directly impacts their performance on these tasks.
\end{itemize}

In conclusion, due to the substantial differences between the two, it is essential to develop specialized approaches for NL2GQL rather than simply adapting Text-to-SQL methods.

\noindent \textbf{Multi-turn Dialogue}

Multi-turn dialogue systems involve an iterative, back-and-forth exchange between a user and a system, where the conversation evolves over multiple turns. These systems aim to refine user queries, explore topics in more depth, and generate contextually appropriate responses based on previous interactions. Unlike single-turn dialogue systems, which address isolated queries, multi-turn dialogues manage dynamic and context-sensitive information flows~\cite{yi2024survey}.

Multi-turn NL2GQL is a specialized form of Multi-turn Dialogue. Unlike other Multi-turn Dialogue systems, NL2GQL focuses on converting natural language into GQL based on a graph database. This distinction makes Multi-turn NL2GQL ideal for dynamic interactions with graph-based data, where each query may involve traversing different paths or nodes. The model must not only understand the current query but also retain information from previous interactions to generate accurate, contextually relevant graph queries. This ability to maintain coherence across multiple turns poses challenges in handling complex graph traversals and evolving contexts.

\noindent \textbf{Multi-turn Knowledge Base Question Answering.}
A knowledge graph is a structured knowledge base represented as a graph, designed to organize vast amounts of real-world information in a flexible and scalable manner. Its primary goal is to enable machines to understand this information and perform reasoning and inference~\cite{zhao2022natural,pan2024unifying}. In contrast, a graph database primarily focuses on efficient data storage and query optimization, rather than on knowledge reasoning and semantic understanding. As such, KBQA emphasizes knowledge-based reasoning and semantic understanding to extract answers from structured knowledge bases, while NL2GQL focuses on constructing effective graph queries.

A typical example of a problem that NL2GQL can solve but KBQA cannot is as follows:

\textbf{Problem:} Find all users who participated in at least two projects in 2023, and whose friends include at least one person from the R\&D department.

\textbf{NL2GQL Solution:} The complex graph traversal logic can be directly expressed using graph query languages like Cypher Pseudo-code:

\begin{verbatim}
MATCH (u:User)-[:PARTICIPATED_IN]->(
p:Project {year: 2023}) 
WITH u, COUNT(p) AS project_count 
WHERE project_count >= 2 
MATCH (u)-[:FRIEND_OF]->(f:User)-
[:BELONGS_TO]->(:Dept {name: "R&D"}) 
RETURN u.name, COLLECT(f.name) 
       AS friends_in_rd
\end{verbatim}

Why KBQA Struggles with This Problem:

\begin{itemize} 
   \item \textbf{Multi-hop Relationship Traversal:}
    This problem requires reasoning across 4 hops: User → Project → Count → Friend → Department. Traditional KBQA systems typically handle only single-hop or fixed-path queries and are not equipped to flexibly manage dynamic path lengths (e.g., recursive traversal of the "FRIEND\_OF" relationship).
    
    \item \textbf{Aggregation and Conditional Combination:}  
    The task involves both an aggregation operation (e.g., COUNT(p) >= 2) and a conditional filter (e.g., friends from the R\&D department). KBQA systems usually cannot combine aggregation functions with multiple entity conditions within the same query.

    \item \textbf{Implicit Logical Dependencies:}  
    The condition "at least one friend belongs to the R\&D department" necessitates an existence check (EXISTS) rather than a simple attribute match. KBQA typically returns explicitly stored triples and cannot dynamically infer such existence conditions.
\end{itemize}

Other NL2GQL-exclusive Capabilities include the following question examples:

\begin{itemize} 
   \item \textbf{Path Queries:}
    Question: “Find the shortest collaboration path from User A to User B, where all nodes in the path are employees who joined after 2020.”

    \textbf{Cypher Pseudo-code}:
    \begin{verbatim}
MATCH (a:User {name: "UserA"}), 
 (b:User {name: "UserB"}),
 path = shortestPath((a)-
 [:COLLABORATES_WITH*]-(b))
WHERE ALL(node IN nodes(path) 
WHERE node:Employee AND 
 node.join_date >= '2020-01-01')
RETURN path
\end{verbatim}
    
    \item \textbf{Dynamic Pattern Reasoning:}  
   Question: “Count the managers in all departments who have more than 10 subordinates and whose subordinates have participated in cross-departmental projects.”

   \textbf{Cypher Pseudo-code}:
    \begin{verbatim}
MATCH (dept:Department)
 <-[:MANAGES]-(manager:Manager)
WITH dept, manager, [(manager)-
 [:MANAGES]->(emp:Employee) | emp] 
 AS subordinates
WHERE size(subordinates) > 10
 AND ANY(emp IN subordinates 
WHERE EXISTS {
    MATCH (emp)-[:PARTICIPATED_IN]
     ->(proj:Project)
    WHERE proj.department 
     <> dept.name
    })
RETURN dept.name AS department, 
 manager.name AS manager, 
 size(subordinates) AS emp_count
\end{verbatim}

   \item \textbf{Temporal Graph Analysis:}  
   Question: “List all stocks that experienced a drop of more than 5\% in a single day after 5 consecutive days of price increases.”

   \textbf{Cypher Pseudo-code}: 

    \begin{verbatim}
MATCH (s:Stock)-[r:HAS_DAILY_DATA]
    ->(d:DailyData)
WITH s, d ORDER BY d.date ASC
WITH s, COLLECT(d) AS data
WHERE size(data) >= 6
  AND ANY(i IN RANGE(0,
    size(data)-6) 
  WHERE 
    REDUCE(isRising = true, 
     j IN [0..4] | 
     isRising AND 
     data[i+j+1].close_price >
      data[i+j].close_price
    ) 
    AND (data[i+5].close_price - 
     data[i+6].close_price) /
     data[i+5].close_price >= 0.05
RETURN s.name AS stock, 
 data[i+5].date AS peak_date, 
 data[i+6].date AS drop_date
\end{verbatim}

\end{itemize}

\begin{algorithm*}[ht!]
\caption{Question Expansion Pattern Selection Algorithm}
\label{alg:pattern_selection}
\KwIn{Set of entities and relations $\{E, R\}$, schema of $G$, set of expansion patterns $\{P1, P2, P3, P4, P5, P6\}$}
\KwOut{Selected expansion pattern and corresponding entities and relations}

\textbf{Step 1: Expansion Pattern Filtering}\\
\For{each expansion pattern $P_i$ in $\{P1, P2, \dots, P6\}$}{
    \If{Pattern $P_i$ meets the predefined conditions based on $E$, $R$, and $G$}{
        Include $P_i$ in the set of valid patterns
    }
    \Else{
        Remove $P_i$ from the set of valid patterns
    }
}

\textbf{Step 2: Expansion Pattern Selection}\\
\For{each valid expansion pattern $P_i$}{
    Set initial weight of $P_i$ as $w(P_i) = \frac{1}{6}$
}
\For{each previously used expansion pattern $P_i$}{
    Halve its weight: $w(P_i) = \frac{w(P_i)}{2}$\\
    Redistribute the halved weight equally among other remaining patterns
}

Select the expansion pattern $P_{\text{selected}}$ with the highest weight: \\
$P_{\text{selected}} = \arg\max_{P_i} w(P_i)$

\textbf{Step 3: Entity and Relation Selection}\\
Determine the potential candidate entities $E_{\text{candidates}}$ based on $P_{\text{selected}}$\\
\For{each candidate entity $e \in E_{\text{candidates}}$}{
    Set initial weight of entity $e$ as $w(e) = \frac{1}{|E_{\text{candidates}}|}$\\
    \If{$e$ has been referenced in the previous dialogue step}{
        Increase $w(e)$ by $\frac{1}{4}$, indicating higher likelihood of selection
    }
    Redistribute the increased weight evenly among other remaining entities
}

Determine the potential relations $R_{\text{candidates}}$ based on $P_{\text{selected}}$\\
\For{each relation $r \in R_{\text{candidates}}$}{
    Assign weight to $r$ using a similar process as entity selection
}

\KwRet{Selected expansion pattern $P_{\text{selected}}$, selected entities, and selected relations}
\end{algorithm*}

\subsection{Question expansion patterns selection algorithm.}
\label{pattern_expansion_algorithm}

In this section, we present our question expansion pattern selection algorithm, a key innovation of this work. As described in Section~\ref{context_manager}, the Context Manager stores a set of entities and relations, along with six expansion patterns.

As illustrated in Algorithm~\ref{alg:pattern_selection}, our algorithm follows three main steps:

\begin{itemize} 
    \item \textbf{Expansion Pattern Filtering:} Based on the set of entities, relations, and the schema of $G$, we sequentially evaluate the conditions for each of the six expansion patterns (P1-P6) using predefined rules. We filter out the patterns that do not meet the necessary conditions.
    \item \textbf{Expansion Pattern Selection:} From the remaining expansion patterns, we select the most appropriate one according to their assigned weights. Initially, each pattern is given a weight of 1/6. If a pattern has already been used, its weight is halved, and the reduced weight is evenly distributed among the other remaining patterns.
    \item \textbf{Entity and Relation Selection:} Once the expansion pattern is selected, we proceed to choose the corresponding entities and relations. In the entity selection process, we first identify the potential candidate entities based on the chosen pattern. Then, we assign weights to these entities. Initially, each potential entity receives an equal weight of 1/|E|, where |E| is the total number of candidate entities. If an entity has been referenced in the previous step of the dialogue, its weight increases by 1/4, indicating a higher likelihood of its selection in the current step. The increased weight is evenly redistributed among the remaining entities to maintain balance. The relation selection follows a similar approach.
\end{itemize}

\subsection{Prompt for Question Generation}
\label{generate_prompt}

As shown in Figure~\ref{fig:data_generate_prompt}, this prompt generates clear and contextually relevant questions based on a schema and dialogue history, following a question expansion pattern. It guides the LLM to produce either an opening question or a follow-up question that incorporates colloquial, informal, and ambiguous expressions to better simulate real user queries, using entity placeholders according to the expansion pattern. The output includes both a raw question with references and a fully disambiguated version, free of placeholders and references, ensuring contextual relevance and structural clarity. It is worth noting that, since we are constructing a Chinese dataset, the prompt is originally written in Chinese. For ease of reading, however, we have provided an English translation.
\begin{figure*}[ht!]
\centering
\resizebox{0.99\textwidth}{!}{
  \begin{tikzpicture}
    \node[draw, rectangle, inner sep=2mm, fill=mylightblue] (rect) {
      \begin{minipage}{\linewidth}
      \textbf{Instruction:}\\
You are an expert in both language processing and NebulaGraph. Given the schema, question expansion pattern, and dialogue history, generate a clear, relevant, and contextually appropriate question by following the rules below:

\begin{enumerate} 
    \item Generate a question based on the schema and dialogue context, ensuring it is contextually relevant and logically continues the conversation. The question should be conversational in style, incorporating ellipses, omissions, and vague expressions wherever appropriate.
    \item Use placeholders for entities, such as: \texttt{[s]} for stock, \texttt{[c]} for chairman, \texttt{[h]} for stockholder, \texttt{[t]} for trade, \texttt{[p]} for public offering fund, \texttt{[f]} for fund manager, \texttt{[i]} for industry, \texttt{[d]} for time, and \texttt{[m]} for numbers.
    \item If the dialogue history is empty, create an opening question. If there is existing dialogue, generate a follow-up question that aligns with the provided question expansion pattern.
    \item Generate the raw question in a conversational style, incorporating relevant references.
    \item Generate the formal question based on the raw question. The formal question should be a disambiguated version of the raw question, clarified and free of placeholders or references. 
\end{enumerate}

\textbf{Input:}\\
\textbf{1. Schema Information:}\\
\{SCHEMA\}\\
\textbf{2. Dialogue History:}\\
\{DIALOGUE\_HISTORY\}\\
\textbf{3. Question Expansion Pattern:}\\
\{QUESTION\_EXPANDING\_PATTERN\}\\

\textbf{Output:}\\
Provide the generated raw question after "Question" and the formal question after "Complete Question" directly.\\

\textbf{Question:}\\

\textbf{Complete Question:}

      \end{minipage}
    };
  \end{tikzpicture}
}
\caption{The prompt for question generation.}
\label{fig:data_generate_prompt}
\end{figure*}

\subsection{Analysis of Dataset Generation Methodology and Dataset Characteristics}
\label{method_dataset_analyse}

\subsubsection{Detailed Mechanisms of Dataset Construction Components}
In order to explain more detailed descriptions of the internal mechanisms of our dataset construction framework components, we provide the following explanations for the key modules: Question Generator, GQL Generator, and GQL Validator and Optimizer.

\paragraph{Question Generator.}  
The Question Generator leverages a LLM to produce contextually coherent questions by conditioning on the dialogue history and relevant schema information. Specifically, the LLM is prompted with both previous turns in the conversation and masked templates to ensure that the generated questions maintain semantic continuity and relevance to the evolving dialogue context. Detailed prompt designs and example outputs are provided in Figure ~\ref{fig:data_generate_prompt}.

\paragraph{GQL Generator.}  
To convert natural language questions into executable GQL commands, the GQL Generator employs a fine-tuned LLM guided by the complete database schema. The generator incorporates the full schema context and uses the reformulated question, which includes disambiguated references and expanded context, to produce accurate and context-aware GQL queries. This approach is inspired by the method described in~\cite{liang2024nat}, which effectively integrates schema constraints to generate GQL.

\paragraph{GQL Validator and Optimizer.}
The GQL Validator and Optimizer modules are responsible for the semantic verification and refinement of generated queries. The Validator executes the generated GQL query against the graph database and compares the results with the expected outcomes inferred from the dialogue context to identify any discrepancies. Upon detecting inconsistencies, the Optimizer uses carefully designed prompts—identical to the refiner prompts described in~\cite{liang2024nat}—to guide the LLM in iteratively revising and improving the query. These prompts emphasize error correction, adherence to the database schema, and maintaining contextual consistency. Further details regarding the prompt design and the iterative optimization process can be found in lines 355–368 of this paper.

Together, these components form a tightly integrated framework that ensures generated questions and GQL queries are both contextually coherent and semantically accurate, thereby effectively supporting the construction of a high-quality multi-turn NL2GQL dataset.

\subsubsection{Effectiveness of Dataset-Based Training for GQL Generation}

The core question raised concerns the ability of current LLMs to generate high-quality multi-turn GQL dialogues, particularly in the absence of task-specific training data. While LLMs such as ChatGPT or Qwen2.5 can generate GQL queries without fine-tuning, the accuracy of such outputs is far from guaranteed. Our framework incorporates a dataset-driven training process to enhance the precision of generated queries and reduce the loss of usable data due to filtering invalid outputs. To date, there exists no more effective method for reliably improving GQL generation quality, especially in complex multi-turn scenarios.

To better understand the effectiveness of our training method and the necessity of filtering, we conducted two additional evaluations:
\begin{itemize}
\item \textbf{(1) Direct generation without filtering:} We generated 1,000 multi-turn dialogue samples without applying any error filtering or training. The results show that the execution accuracy (EX) for single-turn queries was only \textbf{39.8\%}, while the overall multi-turn accuracy (AEX) dropped to just \textbf{8.4\%}. This highlights the poor reliability of direct generation without task-specific fine-tuning or filtering mechanisms
\item \textbf{(2) Fine-tuning with limited data:} We fine-tuned the GQL generator using only 500 annotated samples under the "fine-tuning with all schema" setting and evaluated it on the same benchmark test set as in our main experiments. The resulting execution accuracy (EX) and average execution accuracy (AEX) were \textbf{29.99\%} and \textbf{15.42\%}, respectively—substantially lower than the best-performing results reported in our main paper (EX: \textbf{65.39\%}, AEX: \textbf{38.30\%}). These results further confirm the importance of using a high-quality, sufficiently large training set for accurate GQL generation in multi-turn settings.
\end{itemize}

Moreover, Table~\ref{breakdown_result_round} reveals a dramatic \textbf{50\%} performance drop in both EM and EX scores from Round 1 (R1) to Rounds 5+ (R5+), highlighting that the primary bottleneck lies in maintaining contextual understanding and reasoning across multiple dialogue turns, rather than in single-turn query generation.

These findings suggest that the key limitation is not the dataset itself but rather the inherent difficulty of maintaining dialogue coherence and reasoning across multiple conversational turns. Consequently, targeted dataset design and fine-tuning remain critical components in improving multi-turn GQL generation.

\textbf{It is worth reiterating that directly using LLMs to generate GQL queries often results in low accuracy, far from being satisfactory for practical use. }This necessitates a post-processing pipeline that filters and optimizes the generated GQLs. Our primary goal is to construct a high-quality multi-turn NL2GQL dataset, where maintaining the coherence and scalability of natural language questions is crucial. Given the initially low quality of GQLs produced by the LLM, we apply strict filtering to remove a large portion of erroneous intermediate outputs, thereby ensuring the reliability of the final dataset.

Furthermore, as shown in Table~\ref{table:main_result}, the LLM fine-tuned on our generated dataset significantly outperforms the ICL-based approach across multiple evaluation metrics. This demonstrates that our dataset effectively enhances the LLM’s ability to understand and generate accurate graph queries in multi-turn scenarios.

\subsubsection{Handling of Historical Information in Multi-turn NL2GQL}

In our MTGQL dataset and baseline methods, we explicitly model the interdependency of dialogue history to handle multi-turn queries. Specifically, rather than simply concatenating the entire dialogue sequence, we employ a structured approach in which the dialogue context consists of:

\begin{itemize}
    \item \textbf{Previous questions} — to provide linguistic and semantic context;
    \item \textbf{Previously generated GQL queries} — to preserve formal query structures and constraints;
    \item \textbf{Execution results or answers of prior queries} — to help verify correctness and guide refinements;
    \item \textbf{Entities and relations involved in prior turns} — to focus on relevant schema components.
\end{itemize}

This structured context is maintained and managed by the \textit{Context Manager} module (described in Section 4.2), which reformulates the current user question into a more explicit and self-contained query by referencing the above components. This reformulated question, together with an extracted relevant sub-schema, is then passed to the GQL generation and refinement modules.

We use prompt templates that incorporate these historical elements to guide the language model in generating accurate and context-aware GQL statements. This approach goes beyond naive sequence concatenation by leveraging execution feedback and schema relevance, improving the handling of coreferences, ellipsis, and multi-turn dependencies.

\subsubsection{Keyword Analysis}

To evaluate the richness and syntactic diversity of query expressions in the StockGQL dataset, we conducted a keyword frequency analysis across the training, development, and test sets. Specifically, we focused on core \texttt{nGQL}-related terms, categorized as follows:

\begin{itemize}
    \item \textbf{Query Control:} \texttt{MATCH}, \texttt{GO}, \texttt{FETCH}, \texttt{LOOKUP}, \texttt{WHERE}, \texttt{YIELD}, \texttt{WITH}, \texttt{LIMIT}, \texttt{ORDER BY}, \texttt{GROUP BY}, \texttt{RETURN}
    \item \textbf{Logical Operators:} \texttt{AND}, \texttt{OR}, \texttt{NOT}, \texttt{XOR}
    \item \textbf{Graph Traversal:} \texttt{VERTEX}, \texttt{EDGE}, \texttt{OVER}, \texttt{REVERSELY}, \texttt{BIDIRECT}
    \item \textbf{Aggregation Functions:} \texttt{COUNT}, \texttt{SUM}, \texttt{AVG}, \texttt{MAX}, \texttt{MIN}, \texttt{COLLECT}, \texttt{DISTINCT}
\end{itemize}

Excluding structural keywords such as \texttt{MATCH} and \texttt{RETURN}, which appear in nearly all queries by default, the results in Table~\ref{tab:keyword_statistics} show that each dataset split contains a substantial number of informative and diverse keywords. Notably, the test set contains an average of more than 2.1 such keywords per sample. This reflects the high syntactic complexity and operational diversity of StockGQL, highlighting its effectiveness as a benchmark for evaluating the expressive capabilities of NL2GQL models.

\begin{table}[ht!]
\centering
\renewcommand\arraystretch{1.2}
\setlength{\tabcolsep}{5pt}
\begin{tabular}{@{}lccc@{}}
\toprule
 & Total Keywords &  GQL Count & Avg\\
\midrule
Train & 20479 & 19320 & 1.06 \\
Dev   & 3448 & 3252 & 1.16 \\
Test  & 7352 & 6624 & 1.11 \\
\bottomrule
\end{tabular}
\caption{Statistics of nGQL keyword usage in the StockGQL dataset.}
\label{tab:keyword_statistics}
\vspace{-1em}
\end{table}

\subsubsection{Query Type Statistics}

To better understand the distribution of query intents in the MTGQL dataset, we following the question type categorization framework proposed in~\cite{liang2024aligning}, we conducted a comprehensive statistical analysis of StockGQL. As shown in Table~\ref{query_type_result}, StockGQL covers a diverse range of query types, with particularly high representation in complex categories such as Numerical Sorting, Relationship Filtering, and Relationship Inference.

\begin{table}[ht!]
\centering
\small
\renewcommand\arraystretch{0.8}
\begin{tabular}{lccc}
\toprule
 &  train      & dev  & test   \\ 
\midrule
Entity property         & 2145  & 345   & 770 \\ 
Numerical sorting       & 4039  & 841  & 1448 \\ 
Relationship inference  & 2585  & 415   & 891  \\ 
Yes/No question         & 1281  & 249   & 473  \\ 
Relationship filtering  & 4276  & 602  & 1396 \\ 
Attribute comparison    & 1897  & 274   & 782 \\ 
Edge property           & 1923  & 272   & 635  \\ 
String filtering        & 1174  & 254   & 229\\ 
\bottomrule
\end{tabular}
\caption{\label{query_type_result}
Performance of our method on various types of queries in the FinGQL dataset.}
 \vspace{-1.5em}
\end{table}

\subsection{Expansion Patterns and Alignment with User Behavior}

While it is inherently challenging to ensure that automatically generated questions fully capture the diversity of real user behavior, our goal is to approximate realistic multi-turn interaction scenarios as closely as possible. 

To this end, we define six expansion patterns, each designed to reflect common user intents—such as refining a previous query, shifting focus to related entities, or requesting aggregated information. As shown in Table~\ref{pattern_expand}, these patterns offer structural guidance during data generation. We also include representative examples to illustrate how each pattern constrains and informs the generation of follow-up questions in a multi-turn setting.

Furthermore, as demonstrated in Prompt~\ref{fig:data_generate_prompt}, these patterns are explicitly embedded in the prompt instructions provided to the LLM. We additionally require that the generated questions adopt a conversational tone, featuring ellipses, omissions, and vague expressions where appropriate. These expansion patterns act as soft constraints that help the LLM maintain coherence, contextual relevance, and logical progression across dialogue turns, thereby improving the plausibility and utility of the resulting dataset.

\subsection{Generalization to Other Datasets}

To evaluate the generalization capability of our proposed multi-turn dataset \textbf{MTGQL}, we conducted cross-dataset transfer experiments on \textbf{StockGQL}\cite{liang2024nat}. Specifically, following the method in\cite{liang2024nat}, we fine-tuned the same GQL generator model on MTGQL and directly evaluated it on the StockGQL test set, without any further fine-tuning on StockGQL data.

The results are summarized in Table~\ref{tab:cross_dataset}. We observe that, although the model trained solely on MTGQL does not surpass models directly trained on StockGQL, it still achieves competitive performance, with EM and EX scores exceeding 80%.

Additionally, we explored a joint training strategy where the model was first trained on MTGQL and then fine-tuned on StockGQL. This setting yielded consistent improvements of approximately 4.5\% across all metrics compared to training on StockGQL alone. These results suggest that MTGQL serves as a valuable complementary resource, enhancing the generalization ability and robustness of models for NL2GQL tasks.

\begin{table}[htbp]
  \centering
  
  \begin{tabular}{lcc}
    \toprule
    \textbf{Training Dataset} & \textbf{EM (\%)} & \textbf{EX (\%)} \\
    \midrule
    StockGQL only & 85.44 & 86.25 \\
    MTGQL only    & 81.61 & 80.23 \\
    MTGQL + StockGQL & \textbf{90.15} & \textbf{90.89} \\
    \bottomrule
  \end{tabular}
  \caption{Cross-dataset evaluation: training on MTGQL and testing on StockGQL.}
  \label{tab:cross_dataset}
\end{table}

\subsection{Manual Evaluation Protocol}

To assess the dataset's quality, we conducted a human evaluation involving three domain experts. They independently rated 200 randomly sampled dialogues from each split (train, dev, test), totaling 600 dialogues. Each dialogue was evaluated on four dimensions:

\begin{itemize}
    \item \textbf{Coherence}: logical flow across dialogue turns.
    \item \textbf{Question Diversity}: variety in question types and forms.
    \item \textbf{Coverage}: breadth of entities and relations involved.
    \item \textbf{Semantic Accuracy}: alignment of questions with the schema and their meaningfulness.
\end{itemize}

Each dimension was scored on a 1--5 scale, where 1 = \textit{very poor}, 2 = \textit{poor}, 3 = \textit{fair}, 4 = \textit{good}, and 5 = \textit{excellent}. Detailed guidelines for the scoring are as follows:

\begin{itemize}
    \item \textbf{Coherence}: 
    \begin{itemize}
        \item 1: Dialogue is incoherent or inconsistent. 
        \item 2: Frequent logical gaps. 
        \item 3: Partially coherent with some abrupt transitions. 
        \item 4: Mostly logical and connected. 
        \item 5: Fully coherent and natural dialogue flow.
    \end{itemize}
    
    \item \textbf{Question Diversity}:
    \begin{itemize}
        \item 1: Highly repetitive questions. 
        \item 2: Limited variation in question form or content. 
        \item 3: Moderate diversity. 
        \item 4: Good variation in question types. 
        \item 5: Broad and rich variety of question forms and intents.
    \end{itemize}
    
    \item \textbf{Coverage}: 
    \begin{itemize}
        \item 1: Very narrow focus on one topic or entity. 
        \item 2: Minor variation in entities or relations. 
        \item 3: Involves a few distinct schema elements. 
        \item 4: Covers a range of entity and relation types. 
        \item 5: Broad and comprehensive schema coverage.
    \end{itemize}
    
    \item \textbf{Semantic Accuracy}: 
    \begin{itemize}
        \item 1: Questions are semantically invalid or nonsensical. 
        \item 2: Multiple inconsistencies with schema. 
        \item 3: Generally valid but with minor semantic flaws. 
        \item 4: Mostly correct and meaningful. 
        \item 5: Fully accurate, meaningful, and well-grounded in the schema.
    \end{itemize}
\end{itemize}

Each dialogue was evaluated independently by all three experts, and the final score per dimension was averaged. To ensure consistency in annotation, we computed inter-rater agreement using Cohen’s Kappa, which yielded a score of \textbf{85.76}, indicating a high level of annotation reliability.

\subsubsection{Example Case Analysis}
Here is a sample dialogue excerpt and its evaluation:

\begin{itemize}
    \item \textbf{Dialogue:}
    \begin{itemize}
        \item Q1: ``Who is the CEO of \texttt{[Company A]}?''
        \item Q2: ``What subsidiaries does it own?''
        \item Q3: ``Among them, which were founded after 2010?''
    \end{itemize}
    
    \item \textbf{Expert Scores:}
    \begin{itemize}
        \item \textbf{Coherence: 5} --- Each turn builds naturally on the previous.
        \item \textbf{Diversity: 4} --- Mix of factoid and temporal questions.
        \item \textbf{Coverage: 5} --- Involves various entity types (company, person, subsidiary, time).
        \item \textbf{Semantic Accuracy: 5} --- All questions align well with the schema and are meaningful.
    \end{itemize}
\end{itemize}

% \subsection{Further Analysis of the MTGQL Dataset}
% \label{further_dataset_analyse}

\begin{algorithm*}[ht!]
\caption{Dependency-aware Multi-turn NL2GQL Inference}
\label{alg:da_mt_nl2gql}
\KwIn{Graph database $G$; multi-turn dialogue $C = \{(Q_1, A_1), \dots, (Q_{t-1}, A_{t-1})\}$; current question $Q_t$}
\KwOut{Executable GQL query $GQL_t$}

\textbf{Initialize:} Structured context $\mathcal{H} \leftarrow \emptyset$\;
\For{$i \leftarrow 1$ \KwTo $t-1$}{
    Extract $(Q_i, A_i)$ from $C$\;
    $GQL_i \leftarrow$ previously generated query for $Q_i$\;
    $Entities_i, Relations_i \leftarrow \texttt{Analyze}(GQL_i, A_i)$\;
    $\mathcal{H} \leftarrow \mathcal{H} \cup \{Q_i, GQL_i, A_i, Entities_i, Relations_i\}$\;
}

$Q_t^{\text{explicit}} \leftarrow \texttt{Reformulate}(Q_t, \mathcal{H})$\tcp*[r]{Resolve coreference and ellipsis}

$SubSchema_t \leftarrow \texttt{ExtractRelevantSubSchema}(G, \mathcal{H}, Q_t^{\text{explicit}})$\;

$GQL_t^{\text{init}} \leftarrow \texttt{GQLGenerator}(Q_t^{\text{explicit}}, SubSchema_t)$\;

$A_t^{\text{pred}} \leftarrow \texttt{Execute}(GQL_t^{\text{init}}, G)$\;

\eIf{$\texttt{IsAligned}(A_t^{\text{pred}}, Q_t, \mathcal{H})$}{
    $GQL_t \leftarrow GQL_t^{\text{init}}$\;
}{
    $GQL_t \leftarrow \texttt{Refine}(GQL_t^{\text{init}}, A_t^{\text{pred}}, Q_t, \mathcal{H})$\;
}
\Return{$GQL_t$}
\end{algorithm*}

\subsection{Dependency-aware Method}
\label{da_method}

To address the challenge of modeling multi-turn dependencies in NL2GQL, we propose a \textbf{Dependency-aware Method (DA)}, which extends the Dependency-aware Multi-turn Dataset Construction Framework with necessary adaptations, following the approach of~\cite{liang2024nat} and tailoring it to the MTGQL dataset setting.

The proposed DA method comprises three key components: a \textit{Context Manager}, a \textit{GQL Generator}, and a \textit{GQL Refiner}. These components are designed to collaboratively maintain dialogue coherence, support context-sensitive reasoning, and generate accurate graph queries in multi-turn interactions. The pseudocode of the algorithm is shown in Algorithm~\ref{alg:da_mt_nl2gql}.

\paragraph{Context Manager.} 
This module is responsible for maintaining and organizing the dialogue history across turns. For each turn, it constructs a structured context that includes:
\begin{itemize}
    \item Natural language questions from previous turns;
    \item Corresponding GQL queries generated in earlier turns;
    \item Execution results of those queries;
    \item Involved entities and relations, representing the dynamic subgraph explored so far.
\end{itemize}
Before generating the current turn's query, the Context Manager reformulates the user question into a more explicit, context-independent version. This includes resolving coreferences (e.g., ``their'', ``its'') and filling in ellipses. It also retrieves a relevant sub-schema by identifying schema elements mentioned in both the dialogue history and the current turn, ensuring precise grounding.

\paragraph{GQL Generator.} 
Given the reformulated question and the retrieved sub-schema, this module utilizes a fine-tuned large language model (LLM) to generate a candidate GQL query. Following the method described in~\cite{liang2024nat}, the generator aims to produce structurally and semantically accurate queries aligned with the user's intent in the current dialogue context.

\paragraph{GQL Refiner.}
Due to the inherent difficulty of GQL generation in complex multi-turn settings, we introduce a post-generation refinement step. The Refiner evaluates whether the generated query aligns with the intended meaning of the user input by analyzing its execution result. If inconsistencies are detected, the Refiner prompts the model to revise the query, improving execution correctness and robustness.

\paragraph{Collaboration Mechanism.} 
The three components operate in a tightly coupled workflow. The Context Manager ensures that rich contextual information is provided to the GQL Generator, enabling it to account for prior dialogue turns. The GQL Generator then produces an initial query candidate, which is further validated and refined by the GQL Refiner. This collaborative mechanism ensures continuity, contextual fidelity, and high-quality query generation throughout the multi-turn process.

\vspace{0.5em}
Overall, this dependency-aware pipeline bridges the gap between natural conversational flow and the generation of accurate, executable graph queries, thereby enabling robust and interpretable NL2GQL performance in complex multi-turn scenarios.

\vspace{0.5em}
\begin{table*}[ht!]
\centering
\small

\begin{tabular}{lcccc}
\toprule
\textbf{Error Type} & \textbf{ICL-AS} & \textbf{RSE} & \textbf{FT-AS} & \textbf{DA} \\
\midrule
Schema Selection Errors         & 29\% & 25\% & 27\% & 18\% \\
Contextual Understanding Failures & 37\% & 28\% & 34\% & 21\% \\
Logical Form Generation Errors  & 14\% & 22\% & 19\% & 13\% \\
Ambiguity / Underspecification  & 13\% & 15\% & 12\% & 12\% \\
Execution-based Errors          & 7\%  & 10\% & 8\%  & 6\%  \\
\bottomrule
\end{tabular}
\caption{Distribution of error types among different baseline methods on 300 sampled error cases.}
\label{tab:error_analysis}
\end{table*}

\begin{table*}[ht!]
\centering
\small

\begin{tabular}{ p{6cm}|p{8.5cm} }
\toprule
\textbf{Turn(s) and Prediction} & \textbf{Details and Error Type} \\
\midrule

\textbf{Turn 1:} \textit{Show me the companies invested by Baidu.} \newline
\textbf{Turn 2:} \textit{What about their subsidiaries?} \newline
\textbf{Prediction (ICL-AS):} Returns subsidiaries of all companies. &
Fails to resolve ``their'' as referring to companies invested by Baidu. Contextual history is not retained, leading to incorrect scope. 

\textbf{Error Type:} Contextual Understanding Failure. \\
\midrule

\textbf{Turn:} \textit{Which listed companies are controlled by Tencent and operate in the finance sector?} \newline
\textbf{Prediction (RSE):} Omits ``listed'' constraint. &
Schema extraction covers ``Tencent'' and ``finance sector'', but ``listed'' is ignored in generation due to weak schema grounding.

\textbf{Error Type:} Logical Form Generation Error. \\
\midrule

\textbf{Turn:} \textit{How about its most recent investment?} \newline
\textbf{Prediction (FT-AS):} Returns any investment without ordering. &
Fails to interpret ``most recent'' as temporal ordering. Lacks temporal reasoning or clarification strategy.

\textbf{Error Type:} Ambiguity / Underspecification. \\
\bottomrule

\end{tabular}
\caption{Representative errors and analysis on MTGQL dataset.}
\label{tab:case_analysis}
\end{table*}

\subsection{Further Experimental Results}
\label{further_expr}
\subsubsection{Error Analysis}
To better understand the limitations of our proposed baseline methods on the MTGQL dataset, we conduct a detailed error analysis across the four benchmark baselines: ICL-AS, RSE, FT-AS, and DA. We manually analyze 300 error cases sampled from the test set, categorizing them into distinct failure types inspired by prior analyses in Spider 2.0~\cite{lei2024spider} and adapted to the multi-turn NL2GQL setting.

\paragraph{1. Schema Selection Errors (26\%)}
These errors arise when the model selects incorrect or incomplete schema elements (i.e., node or edge types) for the current turn. This is especially problematic in ICL-AS and FT-AS, which must reason over the entire schema without contextual focus. In multi-turn scenarios, the lack of dynamic schema narrowing often causes confusion, especially when the current utterance implicitly refers to earlier entities.

\paragraph{2. Contextual Understanding Failures (32\%)}
These include failures where the model misunderstands the dependencies between the current utterance and the previous turns. For instance, co-reference resolution (e.g., ``What about its subsidiaries?'') or omitted subject/object references lead to incorrect query generation. While DA performs better by maintaining structured dialogue history, it still suffers in complex chained questions where the dependency is not linear or when entity grounding fails.

\paragraph{3. Logical Form Generation Errors (18\%)}
These involve syntactically valid but semantically incorrect GQL outputs. Common examples include incorrect filtering conditions, missing relation constraints, or reversed edges. The RSE method particularly struggles here when the related schema extraction is too coarse, leading to semantically under-constrained queries.

\paragraph{4. Ambiguity and Underspecification (14\%)}
These errors stem from under-specified questions, where even humans may interpret multiple valid GQLs. For example, ``How about their latest investment?'' may refer to different temporal orders depending on context. Models often make arbitrary choices without proper grounding, especially in ICL-AS where no external clarification mechanism exists.

\paragraph{5. Execution-based Errors (10\%)}
Some errors only become evident after query execution, such as returning empty results due to overly specific filters or semantic mismatches. The DA method mitigates this partially using its GQL Refiner module, but residual issues persist due to imperfect execution feedback alignment.

\paragraph{Summary of Trends}
We observe that multi-turn interaction introduces new challenges absent in single-turn NL2GQL tasks: co-reference resolution, context propagation, and entity linking across turns are key failure points. Baselines relying on static prompts (ICL-AS) or full-schema inputs (FT-AS) tend to suffer from information overload or misalignment. Dependency-aware methods (DA) show promise but remain sensitive to entity tracking and reformulation quality.

\subsubsection{Representative Error Cases}

To further illustrate the limitations of baseline methods on the MTGQL dataset, we present representative error cases, highlighting how multi-turn context and schema interaction contribute to failures.

As shown in Table~\ref{tab:case_analysis}, these representative cases reveal that multi-turn NL2GQL tasks go beyond simple slot-filling. Models must integrate contextual memory, resolve references, and incorporate implicit constraints (e.g., time, status). Current baselines lack robust mechanisms for resolving such ambiguities, motivating future work toward hybrid symbolic-neural architectures or multi-agent dialogue managers.

\end{document}